%% file: naaclhlt2018.tex
\documentclass[11pt,a4paper]{article} 
\usepackage[hyperref]{naaclhlt2018}
\usepackage{times} 
\usepackage{latexsym}
\usepackage{url}
\usepackage{amsfonts}
\usepackage{amsmath}
\usepackage{amssymb}
\usepackage{graphicx}
\usepackage[font={footnotesize}]{caption}
\usepackage{stmaryrd}
\usepackage[shortlabels]{enumitem}
\usepackage{lipsum}
\usepackage{booktabs}

\include{std-macros}

\aclfinalcopy 
\def\aclpaperid{51} 

\newcommand\nl[1]{{\it``#1''}}
\newcommand\den[1]{{\llbracket #1 \rrbracket}}

\newcommand\compwebq{\textsc{ComplexWebQuestions}}
\newcommand\webq{\textsc{WebQuestions}}
\newcommand\webqsp{\textsc{WebQuestionsSP}}
\newcommand\simpqa{\textsc{SimpQA}}
\newcommand\splitqa{\textsc{SplitQA}}
\newcommand\splitqaoracle{\textsc{SplitQAOracle}}
\newcommand\rcqa{\textsc{RCQA}}
\newcommand\splitrcqa{\textsc{SplitRCQA}}
\newcommand\googlebox{\textsc{GoogleBox}}
\newcommand\human{\textsc{Human}}

\newcommand\ignore[1]{}

\title{The Web as a Knowledge-base for Answering Complex Questions}

\author{Alon Talmor \\ Tel-Aviv University \\ {\small \tt alontalmor@mail.tau.ac.il} \\\And Jonathan Berant \\ Tel-Aviv University \\ {\small \tt joberant@cs.tau.ac.il} \\}

\date{}

\begin{document} 
\maketitle 
\begin{abstract} 
Answering complex questions is a time-consuming activity for humans that requires reasoning and integration of information. Recent work on reading comprehension made headway in
answering simple questions, but tackling complex questions is still an ongoing research challenge. Conversely, semantic parsers 
have been successful at handling compositionality, but only when the information resides in a target knowledge-base. In this paper, we present a novel framework for answering broad and complex
questions, assuming answering simple questions is possible using a search engine
and a reading comprehension model. We propose to decompose complex 
questions into a sequence of simple questions, and compute
the final answer from the sequence of answers. 
To illustrate the viability of our approach, we create a new
dataset of complex questions, \compwebq, and present a model that decomposes questions
and interacts with the web to compute an answer.
We empirically demonstrate that question decomposition improves performance from 20.8 precision@1 to 27.5 precision@1 on this new dataset.71
\end{abstract}

\section{Introduction} \label{sec:intro}
Humans often want to answer complex questions that require  reasoning over multiple pieces of evidence, e.g., \nl{From what country is the winner of the Australian Open women's singles 2008?}. Answering such questions in broad domains 
can be quite onerous for humans, because it requires searching and integrating information from multiple sources.

Recently, interest in question answering (QA) has surged in the context of reading
comprehension (RC), where an answer is sought for a question given one or more
documents \cite{hermann2015read,joshi2017triviaqa,rajpurkar2016squad}. Neural models trained over large datasets led to great progress in RC, nearing human-level performance \cite{wang2017gated}. However, analysis of models revealed \cite{jia2017adversarial,chen2016thorough} that they mostly excel at matching questions to local contexts, 
but struggle with questions that require reasoning.
Moreover, RC assumes documents with the  information relevant for the answer are available -- but when questions are complex, even retrieving the documents can be difficult.

Conversely, work on QA through semantic parsing has focused primarily on compositionality: questions are translated to compositional programs that encode a sequence of actions for finding the answer in a knowledge-base (KB) \cite{zelle96geoquery,zettlemoyer05ccg,artzi2013weakly,krishnamurthy2012weakly,kwiatkowski2013scaling,liang11dcs}. However, this reliance on a manually-curated KB has limited the coverage and applicability of semantic parsers.

\FigTop{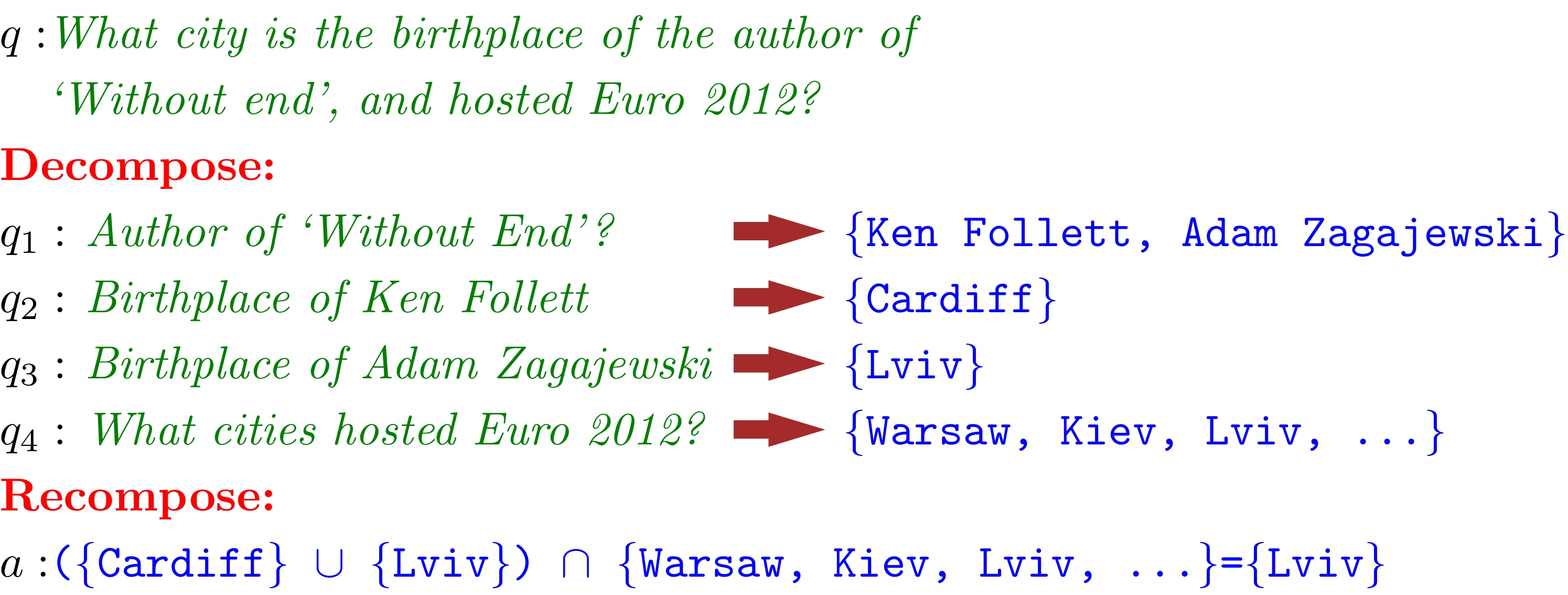}{0.3}{overview}{Given a complex questions $q$, we decompose the question to a sequence of simple questions $q_1, q_2, \dots$, use a search engine and a QA model to answer the simple questions, from which we compute the final answer $a$.}

In this paper we present a framework for QA that is \emph{broad}, i.e., it does not assume information is in a KB or in retrieved documents, and \emph{compositional}, i.e., to compute an answer we must perform some computation or reasoning. Our thesis is that
answering simple questions can be achieved by combining a search engine with a RC model. Thus, answering complex questions can be addressed by 
decomposing the question into a sequence of simple questions, and computing the answer from the corresponding answers. \reffig{overview} illustrates this idea. Our model decomposes the question in the figure into a sequence of simple questions, each is submitted to a search engine, and then an answer is extracted from the search result. Once all answers are gathered, a final answer can be computed using symbolic operations such as union and intersection.


To evaluate our framework we need a dataset of complex questions that calls for reasoning over multiple pieces of information. Because an adequate dataset is missing, we created \compwebq{}, a new dataset for complex questions that
builds on \webqsp{}, 
a dataset that includes pairs of simple questions and their corresponding SPARQL query. We
take SPARQL queries from \webqsp{} and
automatically create more complex queries that include phenomena such as function composition, conjunctions, superlatives and comparatives. Then, we use Amazon Mechanical Turk (AMT) to generate natural language questions, and obtain 
a dataset of 34,689 question-answer pairs (and also SPARQL queries that our model ignores).
Data analysis shows that examples are diverse and that AMT workers perform substantial paraphrasing of the original machine-generated question.

We propose a model for answering complex questions through question decomposition. Our model uses a sequence-to-sequence architecture \cite{sutskever2014sequence} to map utterances to short programs that indicate how to decompose the question and compose the retrieved answers. 
To obtain supervision for our model, we perform a noisy alignment from machine-generated questions to natural language questions and automatically generate noisy supervision for training.\footnote{We differ training from question-answer pairs for future work.}

We evaluate our model on \compwebq and find that  question decomposition
substantially improves precision@1 from 20.8 to 27.5. We find that humans are
able to reach 63.0 precision@1 under a limited time budget, leaving ample room for improvement in future work.

To summarize, our main contributions are:
\begin{enumerate}[nosep]
\item A framework for answering complex questions through question decomposition.
\item A sequence-to-sequence model for question decomposition that substantially improves performance.
\item A dataset of 34,689 examples of complex and broad questions, along with answers, web snippets, and SPARQL queries.
\end{enumerate}
Our dataset, \compwebq{}, can be downloaded from \url{http://nlp.cs.tau.ac.il/compwebq} and our codebase can be downloaded from \url{https://github.com/alontalmor/WebAsKB}.

\section{Problem Formulation} \label{sec:formulation}
Our goal is to learn a model that
given a question $q$ and a black box QA model for answering simple questions, $\textsc{SimpQA}(\cdot)$, 
produces a \emph{computation tree} $t$ (defined below) that decomposes the question and computes the answer. The model is trained from a set of $N$ question-computation tree pairs
$\{q^i, t^i\}_{i=1}^N$ or question-answer pairs $\{q^i, a^i\}_{i=1}^N$.


\FigTop{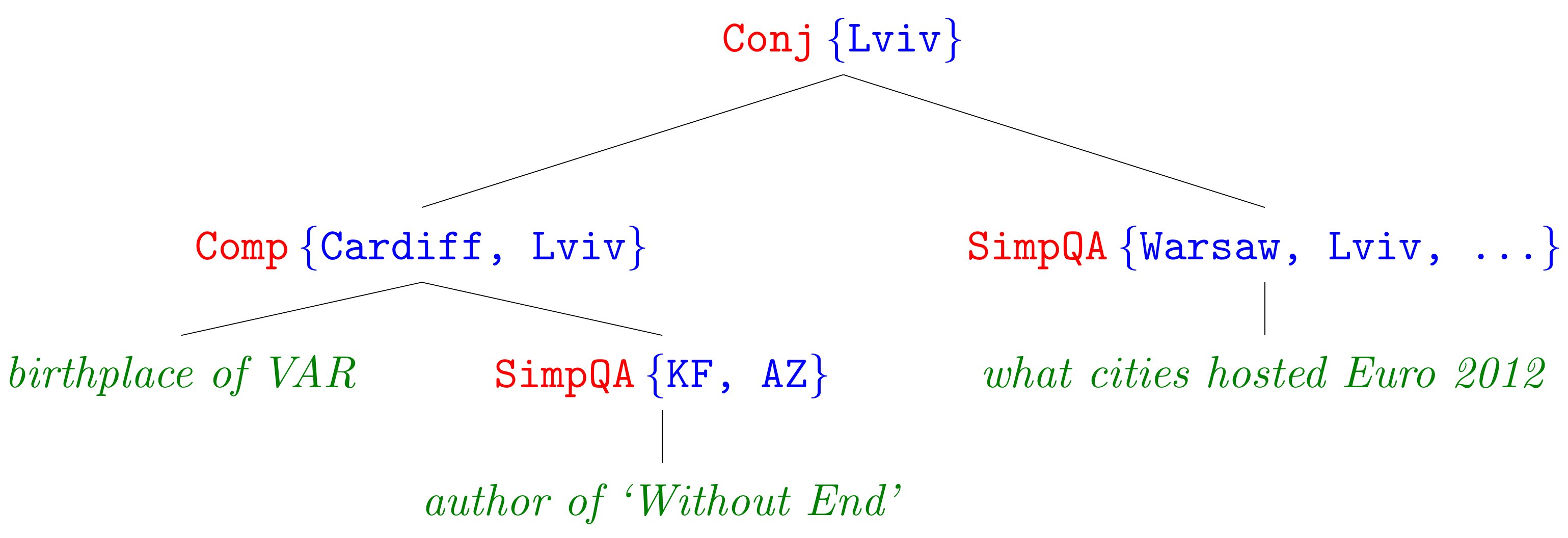}{0.25}{computation_tree}{A  computation tree for \nl{What city is the birthplace of the author of `Without end', and hosted Euro 2012?}. The leaves are strings, and inner nodes are functions (red) applied to their children to produce answers (blue).}

A computation tree is a tree where leaves are labeled with strings, and inner nodes are labeled with functions. The arguments of a function are its children sub-trees. To compute an answer, or \emph{denotation}, from a tree, we recursively apply the function at the root to its children. More formally, given a tree rooted at node $t$, labeled by the function $f$, that has children $c_1(t), \dots, c_k(t)$, the denotation $\den{t} = f(\den{c_1(t)}, \dots, \den{c_k(t)})$ is an arbitrary function applied to the denotations of the root's children. Denotations are computed recursively and the denotation of a string at the leaf is the string itself, i.e., $\den{l} = l$. This is closely related to ``semantic functions"  in semantic parsing \cite{berant2015agenda}, except that we do not interact with a KB, but rather compute directly over the breadth of the  web through a search engine.

\reffig{computation_tree} provides an example computation tree for our running example. Notice that words at the leaves are not necessarily in the original question, e.g., \nl{city} is paraphrased to \nl{cities}.  More broadly, our framework allows paraphrasing questions in any way that is helpful for the function $\textsc{SimpQA}(\cdot)$. Paraphrasing for better interaction with a QA model  has been recently suggested by \newcite{buck2017ask} and \newcite{nogueira2016end}.

We defined the function $\textsc{SimpQA}(\cdot)$ for answering simple questions, but in fact it comprises two components in this work. First, 
the question is submitted to a search engine that retrieves a list of web snippets. Next, a RC model extracts the answer from the snippets. While it is possible to train the RC model jointly with question decomposition, in this work we pre-train it separately, and later treat it as a black box.

The expressivity of our QA model is determined by the functions used, which we turn to next.

\section{Formal Language} \label{sec:language}
Functions in our formal language take arguments and return values that can be strings (when decomposing or re-phrasing the question), sets of strings, or sets of numbers. Our set of functions includes:

\begin{enumerate}[nosep]
\item \textsc{SimpQA$(\cdot)$}:
Model for answering simple questions, which takes a string argument and returns a set of strings or numbers as answer.
\item \textsc{Comp$(\cdot, \cdot)$}:
This function takes a string containing one unique variable \texttt{VAR}, and a set of answers. E.g., in \reffig{computation_tree} the first argument is \nl{birthplace of \texttt{VAR}}, and the second argument is \textsc{``\{Ken Follett, Adam Zagajewski\}"}. The function replaces the variable with each answer string representation and returns their union. Formally, $\textsc{Comp}(q, \sA) = \cup_{a \in \sA} \textsc{SimpQA}(q/a)$, where $q/a$ denotes the string produced when replacing \texttt{VAR} in $q$ with $a$. This is similar to function composition in CCG \cite{steedman00ccg}, or a join operation in $\lambda$-DCS \cite{liang2013lambdadcs}, where the string is a function applied to  previously-computed values.
\item \textsc{Conj}$(\cdot, \cdot)$: 
takes two sets and returns their intersection. Other set operations can be defined analogously. As syntactic sugar, we allow \textsc{Conj}($\cdot$) to take strings as input, which means that we run \textsc{SimpQA}$(\cdot)$ to obtain a set and then perform intersection. The root node in \reffig{computation_tree} illustrates an application of \textsc{Conj}.
\item \textsc{Add}$(\cdot, \cdot)$: 
takes two singleton sets of numbers and returns a set with their addition. Similar functions can be defined analogously. While we support mathematical operations, they were not required in our dataset.
\end{enumerate}

\paragraph{Other logical operations}
In semantic parsing superlative and comparative questions like \nl{What is the highest European mountain?} or \nl{What European mountains are higher than Mont Blanc?} are answered by joining the set of European mountains with their elevation.  While we could add such functions to the formal language, answering such questions from the web is cumbersome: we would have to extract a list of entities and a numerical value for each. Instead, we handle such constructions using \textsc{SimpQA} directly, assuming they are mentioned verbatim on some web document. 

Similarly, negation questions (\nl{What countries are not in the OECD?}) are difficult to handle when working against a search engine only, as this is an open world setup and we do not hold a closed set of countries over which we can perform set subtraction.

In future work, we plan to interface with tables \cite{pasupat2015compositional} and KBs \cite{zhong2017seq2sql}.
This will allow us to perform set operations over well-defined sets, and handle in a compositional manner superlatives and comparatives.

\section{Dataset} \label{sec:dataset}
Evaluating our framework requires a dataset of broad and complex questions that examine the importance of question decomposition. While many QA datasets have been developed recently
\cite{yang2015wikiqa,rajpurkar2016squad,hewlett2016wikireading,nguyen2016ms,onishi2016wdw,hill2015goldilocks,welbl2017constructing}, they lack a focus on the importance of question decomposition.

Most RC datasets contain simple questions that can be answered from a short input document. Recently, \textsc{TriviaQA} \cite{joshi2017triviaqa} presented a larger portion of complex questions, but still most do not require reasoning. Moreover, the focus of \textsc{TriviaQA} is on answer extraction from documents that are given. We, conversely, highlight 
question decomposition for finding the relevant documents. 
Put differently, RC is complementary to question decomposition and can be used as part of the implementation of \textsc{SimpQA}.
In \refsec{experiments} we demonstrate that question decomposition is useful for two different RC approaches.

\subsection{Dataset collection} \label{subsec:collection}
\FigTop{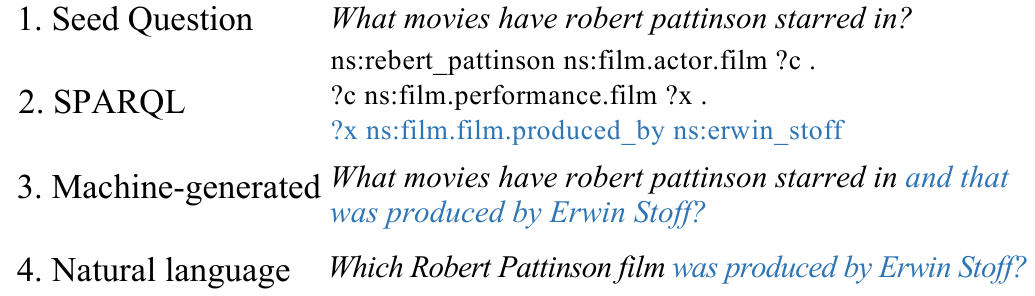}{0.77}{ComplexQuestionGeneration}{Overview of data collection process.}

To generate complex questions we use the dataset \textsc{WebQuestionsSP} \cite{yih2016value}, which contains 4,737 questions paired with SPARQL queries for Freebase \cite{bollacker2008freebase}.
Questions are broad but simple. Thus, we sample question-query pairs, automatically create more complex SPARQL queries, generate automatically questions that are understandable to AMT workers, and then have them paraphrase those into natural language (similar to 
\newcite{wang2015overnight}). We compute answers by executing complex SPARQL queries against Freebase, and obtain broad and complex questions. \reffig{ComplexQuestionGeneration} provides an example for this procedure, and we elaborate next.

\begin{table*}[h]
\begin{center}
\scriptsize{
\begin{tabular}{l|l|l}
 \toprule
 \textbf{Composit.} & \textbf{Complex SPARQL query $r'$}  & \textbf{Example (natural language)} \\ 
 \midrule
\textsc{Conj.} & $r.\ ?x \ \text{pred}_1 \ \text{obj}.$ \textbf{or} & \nl{What films star Taylor Lautner and have costume designs by Nina Proctor?}  \\ 
& $r. \ ?x \ \text{pred}_1 \ ?c. \ ?c \ \text{pred}_2 \ \text{obj}.$  \\
\textsc{Super.} & $r. \ ?x \ \text{pred}_1 \ ?n. \text{ORDER BY DESC}(?n) \ \text{LIMIT} \ 1$ & \nl{Which school that Sir Ernest Rutherford attended has the latest founding date?} \\
\textsc{Compar.} & $r. \ ?x \ \text{pred}_1 ?n. \ \text{FILTER} \ ?n < V$ & \nl{Which of the countries bordering Mexico have an army size of less than 1050?}\\
\textsc{Comp.} & $r[e/y]. \ ?y \ \text{pred}_1 \text{obj}.$ & \nl{Where is the end of the river that originates in Shannon Pot?}\\ 
\toprule
\end{tabular}}
\end{center}
\caption{Rules for generating a complex query $r'$ from a query $r$ ('.' in SPARQL corresponds to logical and). The query $r$ returns the variable $?x$, and contains an entity $e$. We denote by $r[e/y]$ the replacement of the entity $e$ with a variable $?y$. $\text{pred}_1$ and $\text{pred}_2$ are any KB predicates,  $\text{obj}$ is any KB entity, $V$ is a numerical value, and $?c$ is a variable of a CVT type in Freebase which refers to events. The last column provides an example for a NL question for each type.}
\label{tab:sparql_rules}
\end{table*}

\paragraph{Generating SPARQL queries} 
Given a SPARQL query $r$, we create four types of more complex queries: conjunctions, superlatives, comparatives, and compositions.
\reftab{sparql_rules} gives the exact rules for  generation.
For conjunctions, superlatives, and comparatives, we identify queries in \textsc{WebQuestionsSP} whose denotation is a set $\sA, |\sA| \geq 2$, and generate a new query $r'$ whose denotation is a strict subset $\sA', \sA' \subset \sA, \sA' \neq \phi$. For conjunctions this is done by traversing the KB and looking for SPARQL triplets that can be added and will yield a valid set $\sA'$. For comparatives and superlatives we find a numerical property common to all $a \in \sA$, and add a triplet and restrictor to $r$ accordingly. For compositions, we find an entity $e$ in $r$, and replace $e$ with a variable $y$ and add to $r$ a triplet such that the denotation of that triplet is $\{e\}$.

\paragraph{Machine-generated (MG) questions} 
To have AMT workers paraphrase SPARQL queries into natural language, we need to present them in an understandable form. Therefore, we automatically generate a question they can paraphrase. When we generate new SPARQL queries, new predicates are added to the query (\reftab{sparql_rules}). We manually annotated 
687 templates mapping KB predicates to text for different compositionality types (with 462 unique KB predicates), and use those templates to modify the original \text{WebQuestionsSP} question according to the meaning of the generated SPARQL query. 
E.g., the template for  \texttt{\small $?x$ ns:book.author.works\_written \text{obj}} is \nl{the author who wrote OBJ}.
For brevity, we provide the details in the supplementary material.


\paragraph{Question Rephrasing} 
We used AMT workers to paraphrase MG questions into natural language (NL). Each question was paraphrased by one AMT worker and validated by 1-2 other workers. 
To generate diversity, workers got a bonus if the edit distance of a paraphrase was high compared to the MG question. A total of 200 workers were involved, and 34,689 examples were produced with an average cost of 0.11\$ per question. \reftab{sparql_rules} gives an example for each compositionality type.

A drawback of our method for generating data is that because queries are generated automatically the question distribution is artificial from a semantic perspective. Still, 
developing models that are capable of reasoning is an important direction for natural language understanding and \compwebq{}  provides an opportunity to develop and evaluate such models.

To summarize, each of our examples contains a question, an answer, a SPARQL query (that our models ignore), and all web snippets harvested by our model when attempting to answer the question. This renders \compwebq{} useful for both the RC and semantic parsing communities. 

\subsection{Dataset analysis} \label{subsec:data_analysis} 

\compwebq{} builds on the \webq{}  \cite{berant2013freebase}. Questions in \webq{} are usually about properties of entities (\nl{What is the capital of France?}), often with some filter for the semantic type of the answer (\nl{Which director}, \nl{What city}). \webq{} also contains questions that refer to events with multiple entities (\nl{Who did Brad Pitt play in Troy?}). \compwebq{} contains all these semantic phenomena, but we add four compositionality types by generating composition questions (45\% of the times), conjunctions (45\%), superlatives (5\%) and comparatives (5\%).

\paragraph{Paraphrasing} 
To generate rich paraphrases, we gave a bonus to workers that substantially modified MG questions. To check whether this worked, we measured surface similarity between MG and NL questions, and examined the similarity. Using normalized edit-distance and the DICE coefficient, we found that NL questions are different from MG questions and that the similarity distribution has wide support (\reffig{wordsim}).
We also found that AMT workers tend to shorten the MG question (MG avg. length: 16, NL avg. length: 13.18), and use a richer vocabulary (MG \# unique tokens: 9,489, NL \# unique tokens: 14,282).

\FigTop{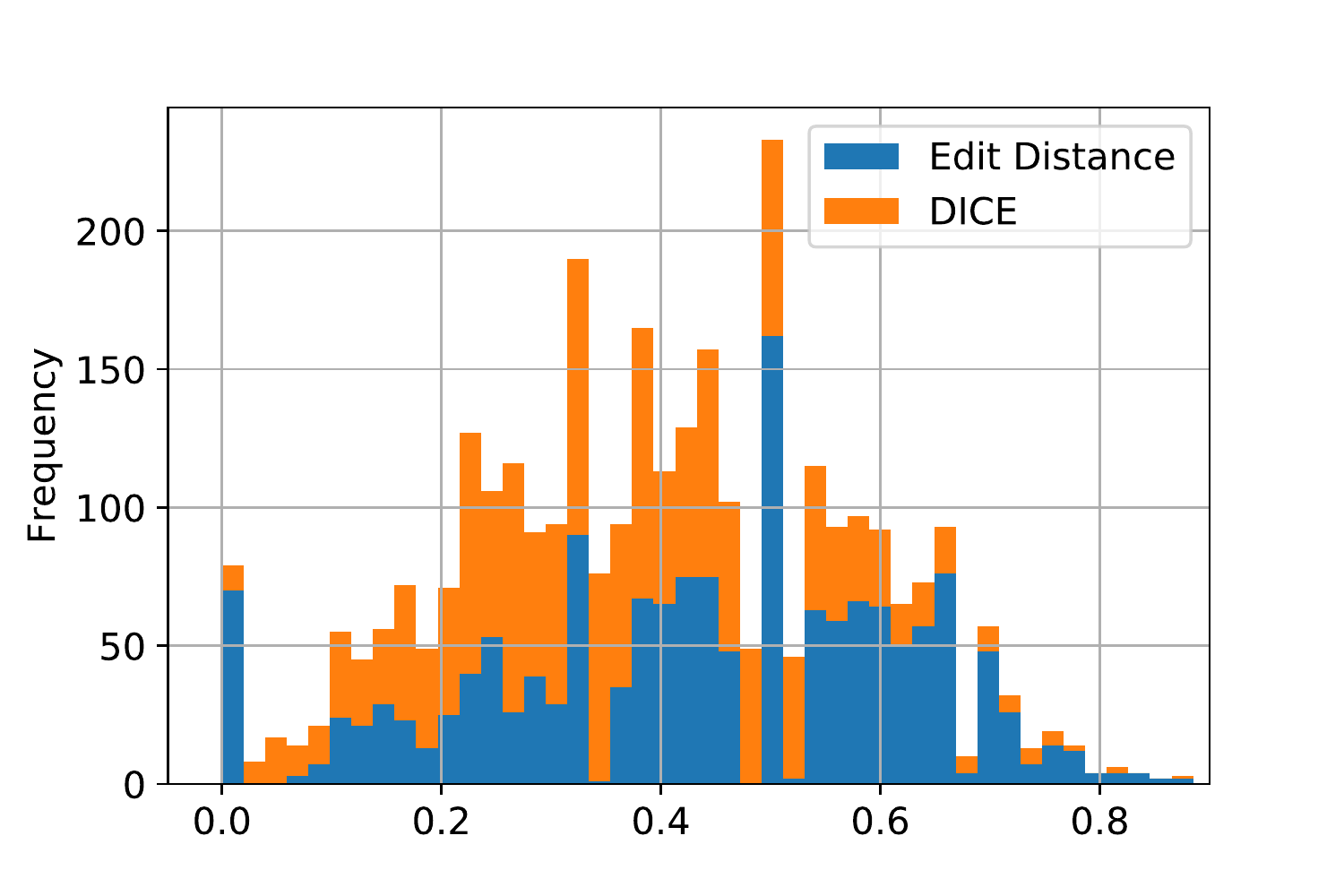}{0.4}{wordsim}{MG and NL questions similarity with normalized edit-distance, and the DICE coefficient (bars are stacked).}

\Fig{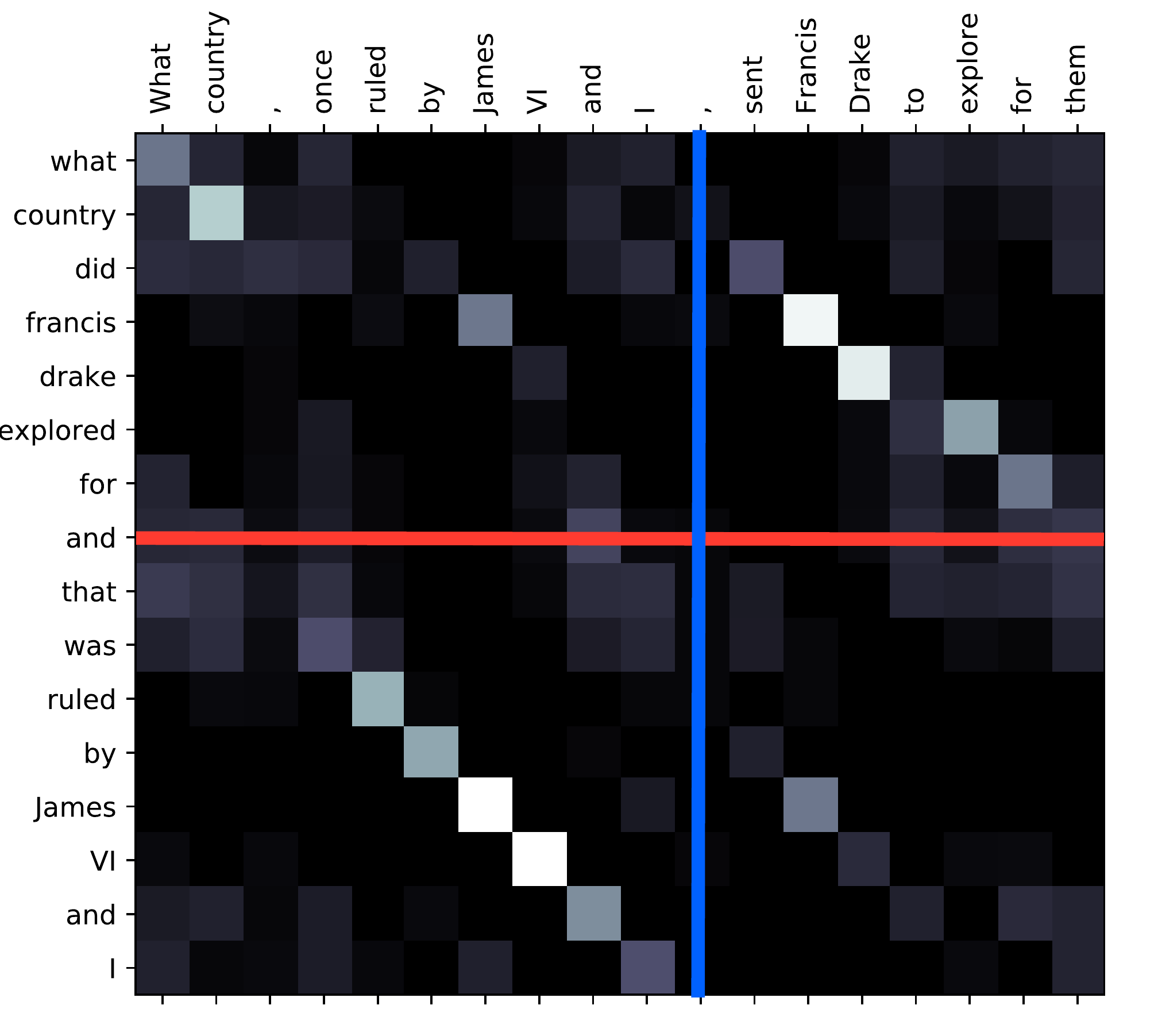}{0.3}{similarity-matrix}{Heat map for similarity matrix between a MG and NL question. The red line indicates a known MG split point. The blue line is the approximated NL split point.}

We created a heuristic for approximating the amount of word re-ordering performed by AMT workers. For every question, we constructed a matrix $A$, where $A_{ij}$ is the similarity between token $i$ in the MG question and token $j$ in the NL question. Similarity is $1$ if lemmas match, or cosine similarity according to GloVe embeddings \cite{pennington2014glove}, when above a threshold, and $0$ otherwise. The matrix $A$ allows us to
estimate whether parts of the MG question were re-ordered when paraphrased to NL
(details in supplementary material). 
We find that in 44.7\% of the conjunction questions and 13.2\%  of the composition questions, word re-ordering happened, illustrating that substantial changes to the MG question have been made. \reffig{similarity-matrix} illustrates the matrix $A$ for a pair of questions with re-ordering.


Last, we find that in \webq{} almost all questions start with a wh-word, but in \compwebq{} 22\% of the questions start with another word, again showing substantial paraphrasing from the original questions.

\paragraph{Qualitative analysis} 
We randomly sampled 100 examples from the development set and manually identified prevalent phenomena in the data.
We present these types in \reftab{rephanalysis} along with their frequency.
In 18\% of the examples a conjunct in the MG question becomes a modifier of a wh-word in the NL question (\textsc{Wh-modifier}). In 22\% substantial word re-ordering of the MG questions occurred, and in 42\% a minor word re-ordering occurred (\nl{number of building floors is 50} paraphrased as \nl{has 50 floors}). AMT workers used a synonym in 54\% of the examples, they omitted words in 27\% of the examples and they added new lexical material in 29\%.

\begin{table*}[t]
\begin{center}
\scriptsize{
\begin{tabular}{l|l|l|l}
 \textbf{Type} & \textbf{MG question} & \textbf{NL question} & {\%}\\ 
 \midrule
\textsc{Wh-modifier}      & \emph{what movies does leo howard play in and that is 113.0 } & \emph{Which Leo Howard movie lasts 113 minutes?} & 18\%  \\ 
 & \emph{minutes long?} & &   \\

\textsc{major reord. }          &  \emph{Where did the actor that played in the film Hancock 2}  &  \emph{What high school did the actor go to who was in the movie Hancock 2?} & 22\% \\ 
 & \emph{go to high school?}& \emph{} & \\

\textsc{minor reord.}          &  \emph{what to do and see in vienna austria} & \emph{What building in Vienna, Austria has 50 floors?} & 42\% \\ 
& \emph{and the number of building floors is 50?}&& \\

\textsc{Synonym}          &  \emph{where does the body of water under Kineshma Bridge start } & \emph{Where does the body of water under Kineshma Bridge originate?} & 54\% \\ 

\textsc{skip word}          &  \emph{what movies did miley cyrus play in and involves } & \emph{What movie featured Miley Cyrus and involved Cirkus? } & 27\% \\ 
& \emph{organization Cirkus?} &  & \\

\textsc{add word}          &  \emph{what to do if you have one day in bangkok and} & \emph{Which amusement park, that happens to be the one that opened} & 29\% \\ 
& \emph{the place is an amusement park that opened earliest?} & \emph{earliest, should you visit if you have only one day to spend in Bangkok?} & \\
\end{tabular}}
\end{center}
\caption{Examples and frequency of prevalent phenomena in the NL questions for a manually analyzed subset (see text).} 
\label{tab:rephanalysis}
\end{table*}

To obtain intuition for operations that will be useful in our model, we analyzed the 100 examples for the types of operations that should be applied to the NL question during question decomposition.
We found that splitting the NL question is insufficient, and that in 53\% of the cases a word in the NL question needs to be copied to multiple questions after decomposition (row 3 in \reftab{rep}). Moreover, words that did not appear in the MG question need to be added in 39\% of the cases, and words need to be deleted in 28\% of the examples. 

\section{Model and Learning} \label{sec:model} 
We would like to develop a model that translates questions into arbitrary computation trees with arbitrary text at the tree leaves. However, this requires training from denotations using methods such as maximum marginal likelihood or reinforcement learning \cite{guu2017bridging} that are difficult to optimize. Moreover, such approaches involve issuing large amounts of queries to a search engine at training time, incurring high costs and slowing down training.

Instead, we develop a simple approach in this paper. We consider a subset of all possible computation trees that allows us to automatically generate noisy full supervision. In what follows, we describe the subset of computation trees considered and their representation, a method for automatically generating noisy supervision, and a pointer network model for decoding.

\begin{table*}[t]
\begin{center}
\scriptsize{
\begin{tabular}{l|l|l}
 \textbf{Program} & \textbf{Question} & \textbf{Split} \\
 \texttt{SimpQA} & \nl{What building in Vienna, Austria has 50 floors} & - \\
 \hline
 \texttt{Comp} $5$ $9$ & \nl{Where is the birthplace of the writer of Standup Shakespeare} &  \nl{Where is the birthplace of VAR} \\
 & & \nl{the writer of Standup Shakespeare} \\
 \hline
 \texttt{Conj} $5$ $1$ & \emph{``What film featured Taylor Swift and} & \nl{What film featured Taylor Swift} \\
 & \emph{ was directed by Deborah Aquila"} & \nl{film and was directed by Deborah Aquila} \\
\hline
\end{tabular}}
\end{center}
\caption{Examples for the types of computation trees that can be decoded by our model.}
\label{tab:rep}
\end{table*}

\paragraph{Representation}
We represent computation trees as a sequence of tokens, and consider trees with at most one compositional operation.  We denote a sequence of question tokens $q_{i:j} = (q_i, \dots, q_j)$, and the decoded sequence by $z$. We consider the following token sequences (see Table \ref{tab:rep}):
\begin{enumerate}[nosep]
\item \texttt{SimpQA}: The function \textsc{SimpQA} is applied to the question $q$ without paraphrasing. 
In prefix notation this is the tree $\textsc{SimpQA}(q)$.
\item \texttt{Comp} $i$ $j$:
This sequence of tokens corresponds to the following computation tree: $\textsc{Comp}(q_{1:i-1} \circ \texttt{VAR} \circ q_{j+1:|q|}, \textsc{SimpQA}(q_{i:j}))$, where $\circ$ is the concatenation operator. This is used for questions where a substring is answered by \textsc{SimpQA} and the answers replace a variable before computing a final answer.
\item \texttt{Conj} $i$ $j$: 
This sequence of tokens corresponds to the computation tree $\textsc{Conj}(\textsc{SimpQA}(q_{0:i-1}), \textsc{SimpQA}(q_j \circ q_{i:|q|}))$. The idea is that conjunction 
can be answered by splitting the question in a single point, where one token is copied to the second part as well (\nl{film} in Table \ref{tab:rep}). If nothing needs to be copied, then $j=-1$.
\end{enumerate}

This representation supports one compositional operation, and a single copying operation is allowed without any re-phrasing. In future work, we plan to develop a more general representation, which will require training from denotations.

\paragraph{Supervision}
Training from denotations is difficult as it involves querying a search engine frequently, which is expensive. Therefore, we take advantage of the
the original SPARQL queries and MG questions
to generate noisy programs for composition and conjunction questions. 
Note that these noisy programs are only used as supervision to avoid the costly process of manual annotation, but the model itself does not assume SPARQL queries in any way.

We generate noisy programs from SPARQL queries in the following manner: First, we automatically identify composition and conjunction questions. 
Because we generated the MG question, we can exactly identify the split points ($i,j$ in composition questions and $i$ in conjunction questions) in the MG question.
Then, we use a rule-based algorithm that 
takes the alignment matrix $A$ (\refsec{dataset}), and approximates the split points in the NL question and the index $j$ to copy in conjunction questions. The red line in \reffig{similarity-matrix} corresponds to the known split point in the MG question, and the blue one is the approximated split point in the NL question. The details of this rule-based algorithm are in the supplementary material.

Thus, we obtain noisy supervision for all composition and conjunction questions and can train a model that translates questions $q$ to representations $z = z_1 \ z_2 \ z_3$, where $z_1 \in \{\texttt{Comp}, \texttt{Conj}\}$ and $z_2, z_3$ are integer indices.

\paragraph{Pointer network}
The representation $z$ points to indices in the input, and thus pointer networks \cite{vinyals2015pointer} are a sensible choice. Because we also need to decode the tokens \textsc{Comp} and \textsc{Conj}, we use ``augmented pointer networks'', \cite{zhong2017seq2sql}:
For every question $q$, an augmented question $\hat{q}$ is created by appending the tokens \textsc{``Comp Conj"} to $q$. This allows us to decode the representation $z$ with one pointer network that at each decoding step points to one token in the augmented question.
We encode $\hat{q}$ with a one-layer GRU \cite{cho2014gru}, and
decode $z$ with a one-layer GRU with attention as  in \newcite{jia2016recombination}. The only difference is that 
we decode tokens from the augmented question $\hat{q}$ rather than from a fixed vocabulary.

We train the model with token-level cross-entropy loss, minimizing $\sum_j \log p_\theta(z_j | x, z_{1:j-1})$. Parameters $\theta$ include the GRU encoder and decoder, and embeddings for unknown tokens (that are not in pre-trained GloVe embeddings \cite{pennington2014glove}).

The trained model decodes \textsc{Comp} and \textsc{Conj} representations, but 
sometimes using $\textsc{SimpQA}(q)$ without decomposition is better. To handle such cases we do the following:
We assume
that we always have access to a score for every answer, provided by the final invocation of \textsc{SimpQA} (in \textsc{Conj} questions this score is the maximum of the scores given by \textsc{SimpQA} for the two conjuncts), and use the following rule to decide if to use the decoded representation $z$ or $\textsc{SimpQA}(q)$. Given the scores for answers given by $z$ and the scores given by $\textsc{SimpQA}(q)$, we return the single answer that has the highest score.
The intuition is that the confidence provided by the scores of $\textsc{SimpQA}$ is correlated with answer correctness. In future work we will train directly from denotations and will handle all logical functions in a uniform manner.


\section{Experiments} \label{sec:experiments}

In this section, we aim to examine whether question decomposition can empirically improve performance of QA models over complex questions.

\paragraph{Experimental setup}
We used 80\% of the examples in \compwebq{} for training, 10\% for development, and 10\% for test, training the pointer network on 24,708 composition and conjunction examples.
The hidden state dimension of the pointer network is $512$, and we used Adagrad \cite{duchi10adagrad} combined with L$_2$ regularization and a dropout rate of $0.25$. We initialize $50$-dimensional word embeddings using GloVe and learn embeddings for missing words. 

\paragraph{Simple QA model} As our \textsc{SimpQA} function, we download the
web-based QA model of \newcite{talmor2017evaluating}. This model sends the
question to Google's search engine and extracts a distribution over answers from
the top-$100$ web snippets using manually-engineered features. We re-train the model on our data with one new feature: for every question $q$ and candidate answer mention in a snippet, we run \textsc{RaSoR}, a RC model by \newcite{lee2016global}, and add the output logit score as a feature. We found 
that combining the web-facing model of \newcite{talmor2017evaluating} and
\textsc{RaSoR}, resulted in improved performance.


\paragraph{Evaluation}
For evaluation, we measure precision@1 (p@1), i.e., whether the highest
scoring answer returned string-matches one of the correct answers (while answers are sets, 70\% of the questions have a single answer, and the average size of the answer set is 2.3).

We evaluate the following models and oracles:
\begin{enumerate}[nosep]
\item \simpqa: running \textsc{SimpQA} on the entire question, i.e., without decomposition.
\item \splitqa: Our main model that answers complex questions by decomposition.
\item \splitqaoracle: An \emph{oracle} model that
chooses whether to perform question decomposition or use \textsc{SimpQA} in hindsight based on what performs better.
\item \rcqa:
This is identical to \simpqa{}, except that we replace the RC model from \newcite{talmor2017evaluating} with the the RC model \textsc{DocQA} \cite{clark2017simple}, whose performance is comparable to state-of-the-art on \textsc{TriviaQA}.
\item \splitrcqa: This is identical to \splitqa{}, except that we replace the RC model from \newcite{talmor2017evaluating} with \textsc{DocQA}.
\item \googlebox: 
We sample $100$ random development set questions and check whether Google returns a box that contains one of the correct answers.
\item \human: 
We sample $100$ random development set questions and manually answer the questions with Google's search engine, including all available information. We limit the amount of time allowed for answering to 4 minutes.
\end{enumerate}

\begin{table}[t]
\begin{center}
\footnotesize{
\begin{tabular}{l|c|c}
 \textbf{System} & \textbf{Dev.} & \textbf{Test} \\
 \hline
 \simpqa & 20.4 & 20.8 \\
 \splitqa & 29.0 & 27.5 \\
 \splitqaoracle & 34.0 & 33.7 \\
 \hline
 \rcqa & 18.7 & 18.6 \\
 \splitrcqa & 21.5 & 22.0\\
 \hline
 \googlebox & 2.5 & - \\
 \human & 63.0 & - \\
  
\end{tabular}}
\end{center}
\caption{precision@1 results on the development set and test set for \compwebq.}
\label{tab:results}
\end{table}

Table~\ref{tab:results} presents the results on the
development and test sets.
\simpqa{}, which does not decompose questions obtained 20.8 p@1, while by
performing question decomposition we substantially improve performance to 27.5 p@1. 
An upper bound with perfect knowledge on when to decompose is given by
\splitqaoracle{} at 33.7 p@1.

\rcqa{} obtained lower performance \simpqa{}, as it was trained on data from a different distribution. More importantly \splitrcqa{} outperforms \rcqa{} 
by 3.4 points, illustrating that this RC
model also benefits from question decomposition, despite the fact that it was not created with question decomposition in mind.
This shows the importance of question decomposition for retrieving documents
from which an RC model can extract answers. 
\googlebox{} finds a correct answer in 2.5\% of the cases, showing that complex questions are challenging for search engines.

To conclude, we demonstrated that question decomposition substantially
improves performance on answering complex questions using two independent RC models.

\paragraph{Analysis}
We estimate human performance (\human{}) at 63.0 p@1. We find that answering complex questions takes roughly 1.3 minutes
on average. For questions we were unable to answer, we found that
in 27\% the answer was correct but exact string match with the gold answers failed;
in 23.1\% the time required to compute the answer was beyond our
capabilities; 
for 15.4\% we could not find an answer on the web;  
11.5\% were of ambiguous nature; 11.5\% involved paraphrasing errors of AMT workers;
and an additional 11.5\% did not contain a correct gold answer.

\splitqa{} decides if to decompose questions or not based on the confidence of
\textsc{SimpQA}. In 61\% of the questions the model chooses to decompose the question,
and in the rest it sends the question as-is to the search engine. 
If one of the strategies (decomposition vs. no decomposition) works, our model
chooses that right one in 86\% of the cases. Moreover, 
in 71\% of these answerable questions, 
only one strategy yields a correct answer.  

 

We evaluate the ability of the pointer network to mimic
our labeling heuristic on the development set. We find that the model outputs
the exact correct output sequence 60.9\% of the time, and allowing errors of
one word to the left and right (this often does not change the final output)
accuracy is at 77.1\%. Token-level accuracy is 83.0\% and allowing
one-word errors 89.7\%. 
This shows that \splitqa{} learned to identify decomposition
points in the questions. We also observed that often \splitqa{}
produced decomposition points that are better than the heuristic, e.g., for \nl{What
is the place of birth for the lyricist of Roman Holiday}, \splitqa{}
produced \nl{the lyricist of Roman Holiday}, but the heuristic produced \nl{the place of birth for the lyricist of Roman Holiday}. Additional examples of \splitqa{} question decompositions are provided in Table~\ref{tab:split}.

\begin{table*}[t]
\begin{center}
\scriptsize{
\begin{tabular}{l|l|l}
 \textbf{Question} & \textbf{Split-1} & \textbf{Split-2} \\
 \emph{``Find the actress who played Hailey Rogers,} & \nl{the actress who played Hailey Rogers} & \nl{Find \texttt{VAR} , what label is she signed to} \\
 \emph{what label is she signed to"} & &  \\
 \hline
 \emph{``What are the colors of the sports team whose } & \emph{``the sports team whose arena stadium} & \nl{What are the colors of \texttt{VAR}} \\
 \emph{arena stadium is the AT\&T Stadium"} & \emph{is the AT\&T Stadium"} &  \\
 \hline
 \emph{``What amusement park is located in Madrid } & \emph{``What amusement park is located in } & \nl{park includes the stunt fall ride} \\
 \emph{Spain and includes the stunt fall ride"} &  \emph{Madrid Spain and"}&  \\
 \hline
 \emph{``Which university whose mascot is } & \emph{``Which university whose mascot is } & \nl{university Derek Fisher attend} \\
 \emph{The Trojan did Derek Fisher attend"} & \emph{The Trojan did"} &  \\
 \hline
\end{tabular}}
\end{center}
\caption{Examples for question decompositions from \splitqa{}.}
\label{tab:split}
\end{table*}


\paragraph{ComplexQuestions}
To further examine the ability of web-based QA models, we run an experiment against \textsc{ComplexQuestions} \cite{bao2016constraint}, a small 
dataset of question-answer pairs  
designed for semantic
parsing against Freebase.

\begin{table}[t]
\begin{center}
\scriptsize{
\begin{tabular}{l|c|c}
 \textbf{System} & \textbf{Dev. F$_1$} & \textbf{Test F$_1$} \\
 \hline
 \simpqa & 40.7  & 38.6 \\
 \textsc{SplitQARule} & 43.1  &  39.7 \\ 
 \textsc{SplitQARule++} & 46.9  & - \\
 \hline
 \textsc{CompQ} & - & \textbf{40.9} \\
\end{tabular}}
\end{center}
\caption{F$_1$ results for \textsc{ComplexQuestions}.}
\label{tab:compquestions}
\end{table}

We ran \simpqa{} on this dataset (\reftab{compquestions}) and obtained 38.6 F$_1$
(the official metric), 
slightly lower than
\textsc{CompQ}, the best system, which operates directly against
Freebase.
\footnote{By adding the output logit from \textsc{RaSor}, we improved test F$_1$ from 32.6, as reported by \newcite{talmor2017evaluating}, to 38.6.} By analyzing the training data, we found that we can
decompose \textsc{Comp} questions with a rule  that splits the question when the words
\nl{when} or \nl{during} appear, e.g., \nl{Who was vice president when JFK was president?}.\footnote{The data is too small to train our decomposition model.} We decomposed questions with this rule and obtained 39.7 F$_1$ (\textsc{SplitQARule}).
Analyzing the development set errors, we found that occasionally
\textsc{SplitQARule} returns a correct answer that fails to string-match with
the gold answer. By manually fixing these cases, our development set F$_1$ reaches 46.9 (\textsc{SplitQARule++}).
Note that \textsc{CompQ} does not suffer from any string matching issue, as it operates directly against the Freebase KB and thus is guaranteed to output the answer in the correct form.
This short experiment shows that a web-based QA model can rival a semantic parser that works
against a KB, and that simple question decomposition is beneficial and leads to results comparable to state-of-the-art.



\section{Related work} \label{sec:related} 
This work is related to a body of work in semantic parsing and RC, in particular to  datasets that focus on complex
questions such as \textsc{TriviaQA} \cite{joshi2017triviaqa},
\textsc{WikiHop} \cite{welbl2017constructing} and \textsc{Race}
\cite{lai2017race}. Our distinction is in proposing a framework for complex QA
that focuses on question decomposition.

Our work is related to \newcite{chen2017reading} and
\newcite{watanabe2017question}, who combined retrieval and answer extraction
on a large set of documents. We work against the entire web, and propose
question decomposition for finding information.

This work is also closely related to \newcite{dunn2017searchqa} and 
\newcite{buck2017ask}: we start with questions directly and do not assume documents are given.
\newcite{buck2017ask} also learn to phrase questions given a black box QA model,
but while they focus on paraphrasing, we address 
decomposition.
Using a black box QA model is challenging because you can not assume differentiability, and reproducibility is difficult as black boxes change over time. Nevertheless, we argue that such QA setups provide a holistic view to the problem of QA and can shed light on important research directions going forward.

Another important related research direction is \newcite{iyyer2016answering}, who answered complex questions by decomposing them. However, they used crowdsourcing to obtain direct supervision for the gold decomposition, while we do not assume such supervision. Moreover, they work against web tables, while we interact with a search engine against the entire web.

\section{Conclusion} \label{sec:conclusion}
In this paper we propose a new framework for answering complex questions that is
based on question decomposition and interaction with the web. We develop a model under this framework and demonstrate it improves
complex QA performance on two datasets and using two RC models.
We also
release a new dataset, \compwebq{}, including questions, SPARQL programs,
answers, and web snippets harvested by our model. We believe this dataset
will serve the QA and semantic parsing communities, drive research
on compositionality, and push the community to work on holistic solutions for QA.

In future work, we plan to train our model directly from weak supervision, i.e., denotations, and to extract information not only from the web, but also from structured information sources such as web tables and KBs.

\section*{Acknowledgements} We thank Jonatahn Herzig, Ni Lao, and the anonymous reviewers
for their constructive feedback. This
work was supported by 
the Samsung runway project and 
the Israel Science
Foundation, grant 942/16.

\bibliography{all}
\bibliographystyle{acl_natbib}
\newpage
\section*{Supplementary Material}
\section*{Dataset} \label{sec:dataset}

\paragraph{Generating SPARQL queries} 
Given a SPARQL query $r$, we create four types of more complex queries: conjunctions, superlatives, comparatives, and compositions. For conjunctions, superlatives, and comparatives, we identify SPARQL queries in \textsc{WebQuestionsSP} whose denotation is a set $\sA, |\sA| \geq 2$, and generate a new query $r'$ whose denotation is a strict subset $\sA', \sA' \subset \sA, \sA' \neq \phi$.   We also discard questions that contain the answer within the new machine-generated questions.

For conjunctions this is done by traversing the KB and looking for SPARQL triplets that can be added and will yield a valid set $\sA'$.  

For comparatives and superlatives this is done by finding a numerical property common to all $a \in \sA$, and adding a clause to $r$ accordingly.  

For compositions, we find an entity $e$ in $r$, and replace $e$ with a variable $y$ and add to $r$ a clause such that the denotation of the clause is $\{e\}$.  We also check for discard ambiguous questions that yield more than one answer for entity $e$.

\begin{table*}[h]
\begin{center}
\scriptsize{
\begin{tabular}{l|l|l}
 \toprule
 \textbf{Composit.} & \textbf{Complex SPARQL query $r'$}  & \textbf{Example (natural language)} \\ 
 \midrule
\textsc{Conj.} & $r.\ ?x \ \text{pred}_1 \ \text{obj}.$ \textbf{or} & \nl{What films star Taylor Lautner and have costume designs by Nina Proctor?}  \\ 
& $r. \ ?x \ \text{pred}_1 \ ?c. \ ?c \ \text{pred}_2 \ \text{obj}.$  \\
\textsc{Super.} & $r. \ ?x \ \text{pred}_1 \ ?n. \text{ORDER BY DESC}(?n) \ \text{LIMIT} \ 1$ & \nl{Which school that Sir Ernest Rutherford attended has the latest founding date?} \\
\textsc{Compar.} & $r. \ ?x \ \text{pred}_1 ?n. \ \text{FILTER} \ ?n < V$ & \nl{Which of the countries bordering Mexico have an army size of less than 1050?}\\
\textsc{Comp.} & $r[e/y]. \ ?y \ \text{pred}_1 \text{obj}.$ & \nl{Where is the end of the river that originates in Shannon Pot?}\\ 
\toprule
\end{tabular}}
\end{center}
\caption{Rules for generating a complex query $r'$ from a query $r$ ('.' in SPARQL corresponds to logical and). The query $r$ returns the variable $?x$, and contains an entity $e$. We denote by $r[e/y]$ the replacement of the entity $e$ with a variable $?y$. $\text{pred}_1$ and $\text{pred}_2$ are any KB predicates,  $\text{obj}$ is any KB entity, $V$ is a numerical value, and $?c$ is a variable of a CVT type in Freebase which refers to events. The last column provides an example for a NL question for each type.}
\label{tab:sparql_rules}
\end{table*}

\reftab{sparql_rules} gives the exact rules for  generation.

\paragraph{Machine-generated (MG) questions} 
To have AMT workers paraphrase SPARQL queries into natural language, we need to present them in an understandable form. Therefore, we automatically generate a question they can paraphrase. When we generate SPARQL queries, new predicates are added to the query (\reftab{sparql_rules}). We manually annotate 
503 templates mapping predicates to text for different compositionality types (with 377 unique KB predicates). 
We annotate the templates in the context of several machine-generated questions to ensure that they result 
templates are in understandable language.

We use those templates to modify the original \textsc{WebQuestionsSP} question according to the meaning of the generated SPARQL query. E.g., the template for $?x \texttt{ ns:book.author.works\_written \text{obj}}$ is \nl{the author who wrote OBJ}.
\reftab{templates} shows various examples of such templates. \nl{Obj} is replaced in turn by the actual name according to Freebase of the object at hand.
Freebase represents events that contain multiple arguments using a special node in the knowledge-base called CVT that represents the event, and is connected with edges to all event arguments.
Therefore, some of our templates include two predicates that go through a CVT node, and they are denoted in \reftab{templates} with '+'.     

To fuse the templates with the original \textsc{WebQuestionsSP} natural language questions, templates contain lexical material that glues them back to the question conditioned on the compositionality type.
For example, in \textsc{Conj} questions we use the coordinating phrase \nl{and is},
so that \nl{the author who wrote OBJ} will produce \nl{Who was born in London and is the author who wrote OBJ}.

\FigTop{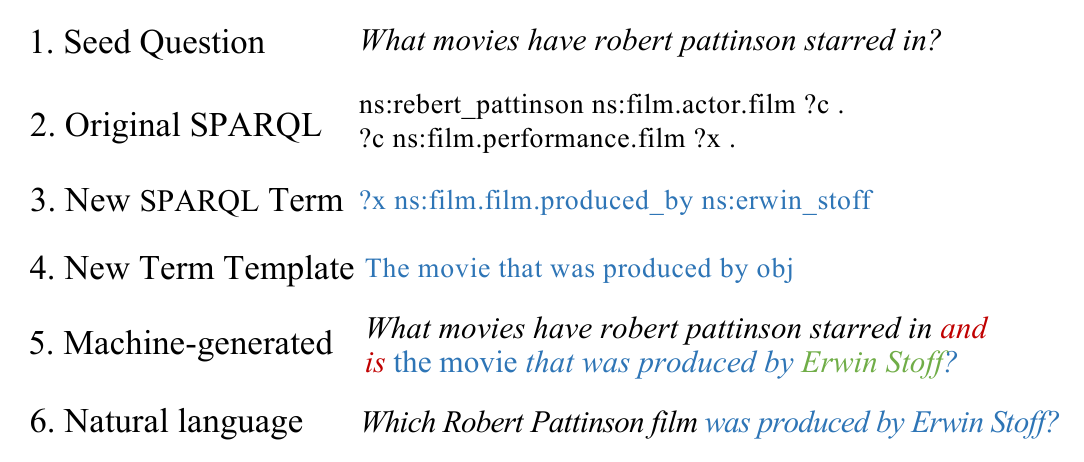}{0.77}{ComplexQuestionGeneration}{Overview of data collection process. Blue text denotes different stages of the term addition, green represents the obj value, and red the intermediate text to connect the new term and seed question }

\begin{table*}[t]
\begin{center}
\scriptsize{
\begin{tabular}{l|l}
  \textbf{Freebase Predicate} & \textbf{Template} \\
  ns:book.author.works\_written &  \nl{the author who wrote obj} \\

ns:aviation.airport.airlines + ns:aviation.airline\_airport\_presence.airline &  \nl{the airport with the obj airline} \\

ns:award.competitor.competitions\_won &  \nl{the winner of obj} \\

ns:film.actor.film + ns:film.performance.film &  \nl{the actor that played in the film obj} \\
\end{tabular}}
\end{center}
\caption{Template Examples}
\label{tab:templates}
\end{table*}

\paragraph{First word distribution}
\Fig{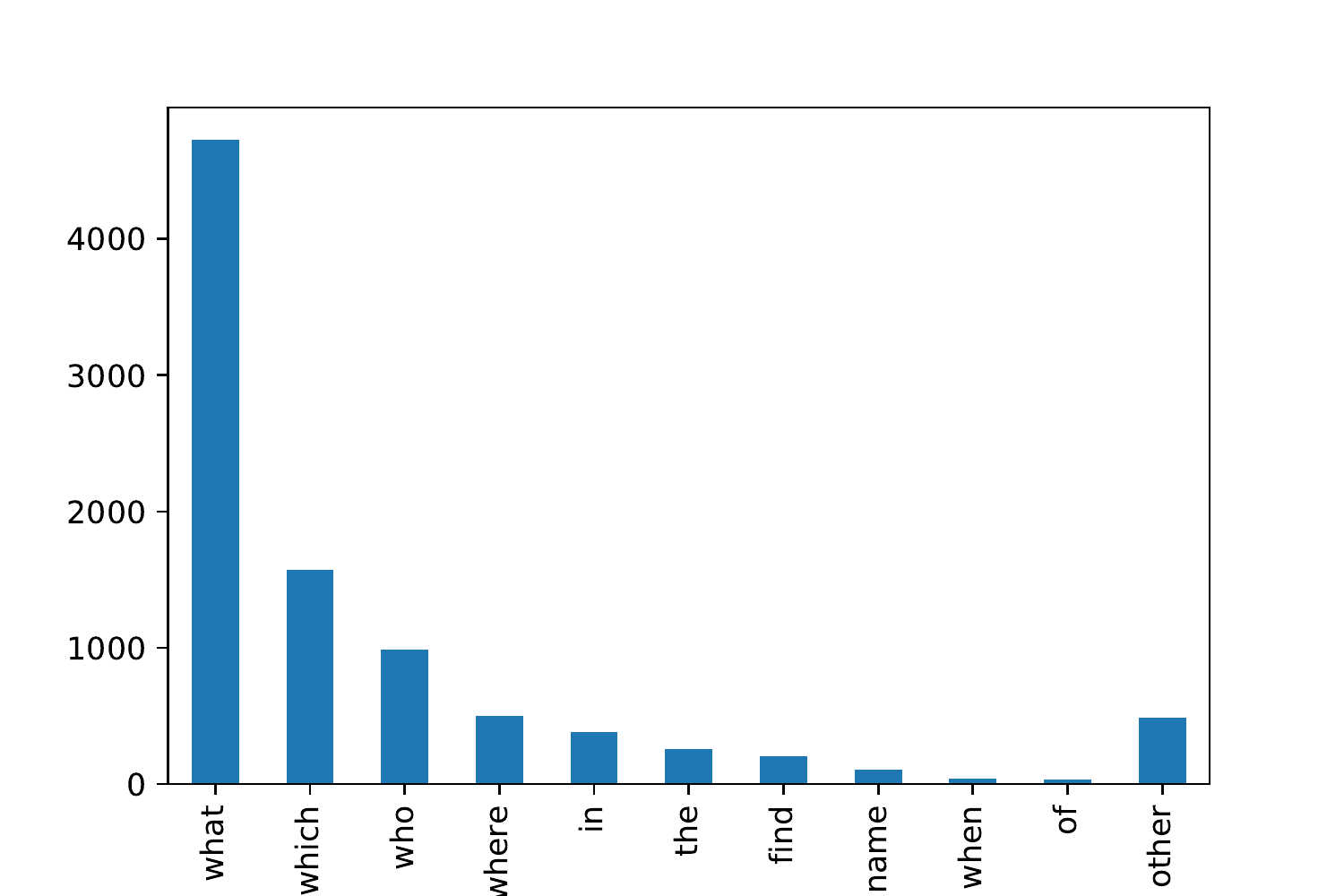}{0.55}{first-word}{First word in question distribution}\label{fig:first-word-distrubition}

We find that in \webq{} almost all questions start with a wh-word, but in \compwebq{} 22\% of the questions start with another word, again showing substantial paraphrasing from the original questions. \reffig{first-word} Shows the distribution of first words in questions.

\section*{Generating noisy supervision} \label{sec:model}  

We created a heuristic for approximating the amount of global word re-ordering performed by AMT workers and creating noisy supervision. For every question, we constructed a matrix $A$, where $A_{ij}$ is the similarity between token $i$ in the MG question and token $j$ in the NL question. Similarity is $1$ if lemmas match, or the cosine distance according to GloVe embedding, when above a threshold, and $0$ otherwise. This allows us to compute an approximate word alignment between the MG question and the NL question tokens  and assess whether word re-ordering occurred.

For a natural language \textsc{Conj} question of length $n$ and a machine-generated question of length $m$ with a known split point index $r$, the algorithm first computes the best point to split the NL question assuming there is no re-ordering. This is done iterating over all candidate split points $p$, and returning the split point $p^{*}_1$ that maximizes:
\begin{equation}\label{eq:1}
\sum_{\substack{
   0 \leq i<p
  }} 
 \max_{0 \leq j<r}
 {A(i,j)}
+ 
\sum_{\substack{
   p \leq i<n
  }} 
 \max_{r \leq j<m}
 {A(i,j)}
\end{equation}

We then compute $p^{*}_1$ by trying to find the best split point, assuming that there is re-ordering in the NL questions: 

\begin{equation}\label{eq:2}
\sum_{\substack{
   0 \leq i<p
  }} 
 \max_{r \leq j<m}
 {A(i,j)}
+ 
\sum_{\substack{
   p \leq i<n
  }} 
 \max_{0 \leq j<r}
 {A(i,j)}
\end{equation}

We then determine the final split point and whether re-ordering occurred by comparing the two values and using the higher one.

In \textsc{Comp} questions,  two split points are returned, representing the beginning and end of the phrase that is to be sent to the QA model.
Therefore, if $r_1, r_2$ are the known split points in the machine-generated questions, we return $p_1, p_2$ that maximize: 
\begin{align*}\label{eq:3}
\begin{split}
&\sum_{\substack{
   0 \leq i<p_1
  }} 
 \max_{0 \leq j<r_1}
 {A(i,j)}
+ 
\sum_{\substack{
   p_1 \leq i<p_2
  }} 
 \max_{r_1 \leq j<r_2}
 {A(i,j)}
 \\
&+ \sum_{\substack{
   p_2 \leq i<n
  }} 
 \max_{r_2 \leq j<m}
 {A(i,j)}.
\end{split}
\end{align*}

\reffig{similarity-matrix} illustrates finding the split point for a \textsc{Conj} questions by using equation \eqref{eq:2}. The red line in \reffig{similarity-matrix} corresponds to the known split point in the MG question, and the blue one is the estimated split point $p^*$ in the NL question.

\Fig{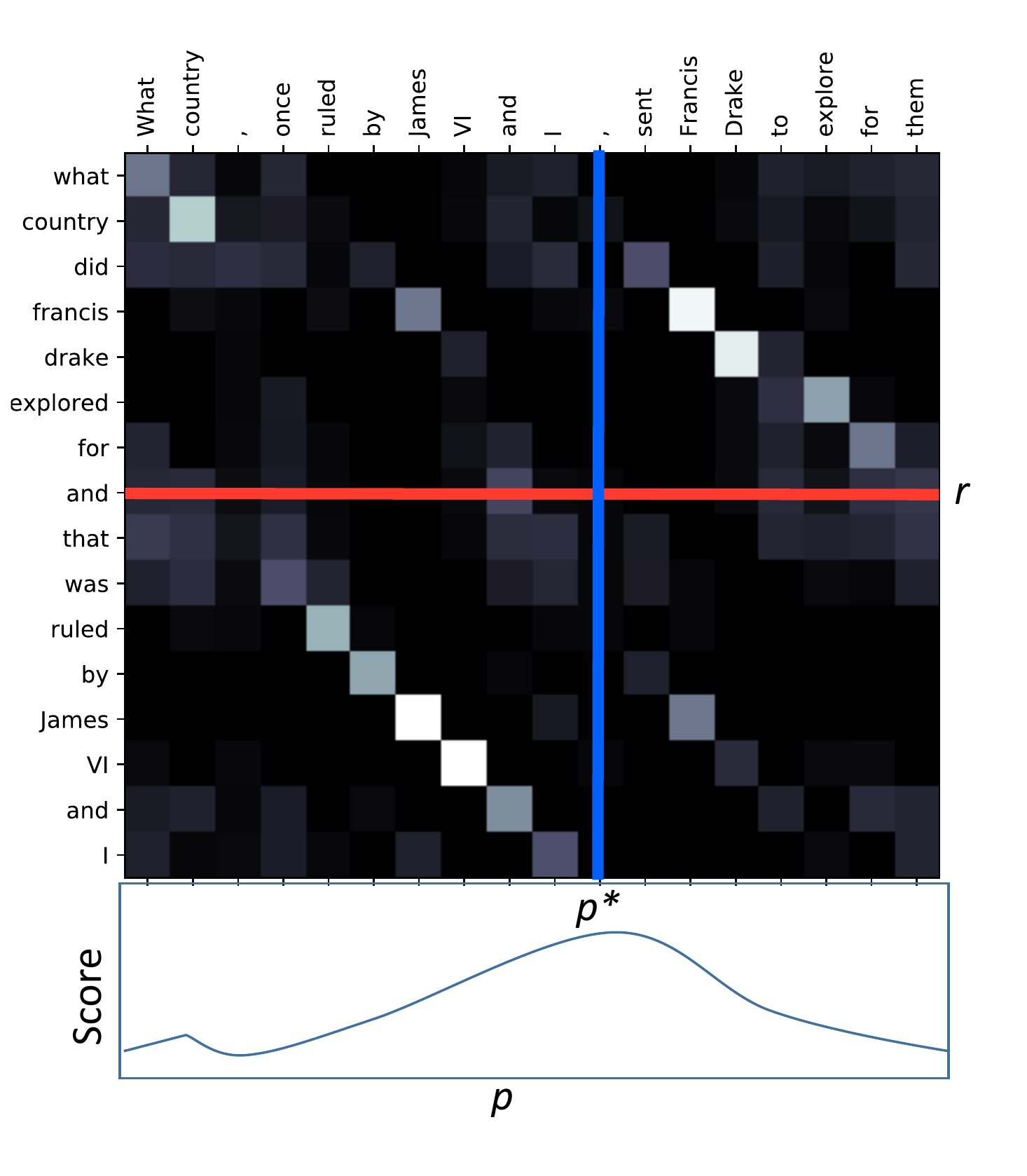}{0.55}{similarity-matrix}{Heat map for similarity matrix between an MG and NL question. The red line indicates a known MG split point. The blue line is the approximated NL split point. Below is a graph of each candidate split point score.}\label{fig:similarity-matrix}
\end{document}

%% file: std-macros.tex
\newcommand\sA{\ensuremath{\mathcal{A}}}

\newcommand\Fig[4]{\begin{figure}[ht] \begin{center} \includegraphics[scale=#2]{#1} \end{center} \caption{\label{fig:#3} #4} \end{figure}}

\newcommand\FigTop[4]{\begin{figure}[t] \begin{center} \includegraphics[scale=#2]{#1} \end{center} \caption{\label{fig:#3} #4} \end{figure}}

\newcommand\refsec[1]{Section~\ref{sec:#1}}

\newcommand\reffig[1]{Figure~\ref{fig:#1}}

\newcommand\reftab[1]{Table~\ref{tab:#1}}
